\documentclass[runningheads]{llncs}

\usepackage[table,xcdraw]{xcolor}
\usepackage{graphicx}
\usepackage{array}
\usepackage{graphicx}
\usepackage{multirow}
\usepackage{ragged2e}
\usepackage{colortbl}
\usepackage{booktabs}
\usepackage{svg}
\usepackage[table]{xcolor}
\definecolor{mycolor1}{HTML}{7ED954}
\definecolor{mycolor2}{HTML}{99e17a}
\definecolor{mycolor3}{HTML}{bfecac}
\definecolor{mycolor4}{HTML}{d9f4cd}
\usepackage{pifont}
\usepackage{eccv}

\usepackage{eccvabbrv}

\usepackage{graphicx}
\usepackage{booktabs}

\usepackage[accsupp]{axessibility}  % Improves PDF readability for those with disabilities.

\makeatletter
\newcommand{\printfnsymbol}[1]{\textsuperscript{\@fnsymbol{#1}}}
\makeatother
\usepackage{hyperref}

\usepackage{orcidlink}

\begin{document}

\title{CROMPT: Adapting CLIP via Cross-Modal Prompting for Robust Zero-Shot Generalization under Adverse Weather} 

\titlerunning{CROMPT}

\author{
Aniruddh Sikdar\inst{1}\thanks{These authors contributed equally to this work.} \and
Arya Kishor\inst{1} * \and
Snehashis Dhibar\inst{2} \and
Suresh Sundaram\inst{1,2}
}

\authorrunning{A. Sikdar et al.}

\institute{
Department of Aerospace Engineering, Indian Institute of Science, Bengaluru, India
\and
Robert Bosch Centre for Cyber Physical Systems, Indian Institute of Science, Bengaluru, India
}

\maketitle

\begin{abstract}
Domain adaptation for semantic segmentation is challenging when target-domain images are unavailable during training. Zero-shot domain adaptation addresses this by adapting models using only textual descriptions of the target domain. However, existing prompt-learning methods are not well suited for ResNet-based CLIP vision backbones or pixel-level cross-modal feature alignment. We propose \textbf{CROMPT}, a prompt-learning framework for zero-shot domain adaptation in semantic segmentation. CROMPT introduces \textbf{Cross-Modal Feature Augmentation (CROMA)}, combining Random Visual Prompting with \textbf{Hybrid Textual Prompting (HTP)} to jointly adapt visual and textual representations. We further propose \textbf{CROMA-VL}, which leverages vision--language models to generate domain-conditioned prompts, and a progressive easy-to-hard adaptation strategy for improved domain invariance. Experiments on two benchmark datasets demonstrate consistent improvements over strong baselines and state-of-the-art methods, achieving \textbf{1.63\%--2.63\% mIoU} gains while maintaining inference-time complexity. These results establish CROMPT as an effective and efficient framework for zero-shot multi-domain semantic segmentation.

\end{abstract}

%%%%%%%%%%%%%%%%%%%%%%%%%%%%%%%%%%%%%%%%%%%%%%%%%%%%%%%%%%%%%%%%%%%%%%%%%%%%%%%%
\section{Introduction}

Computer vision systems based on deep learning, widely used in robotic platforms such as autonomous vehicles \cite{ chattopadhyay2023pasta} and mobile robots, must achieve strong generalization to remain robust across varying illumination, weather patterns, and geographical environments.  Nevertheless, when exposed to out-of-distribution data, their performance often degrades significantly, which constrains their reliability in complex and unpredictable real-world environments \cite{ovadia2019can}.
Unsupervised Domain Adaptation (UDA) mitigates domain discrepancies by learning transferable, domain-invariant representations that bridge a labeled source domain and an unlabeled target domain. In real-world settings, however, access to target data is often restricted due to privacy concerns, security regulations, financial costs, or limited data availability. Moreover, even when such data can be obtained, it may not fully reflect the wide range of variations encountered in practice, thereby constraining the scalability and robustness of UDA methods.

Recent advances in vision–language foundation models have greatly improved image–text alignment. CLIP-based encoders, trained on 400 million web-scale image–text pairs, exemplify this progress, offering strong adaptability, open-set recognition, and cross-domain transfer capabilities, often outperforming vision-only backbones. Consequently, many studies leverage the zero-shot potential of vision–language models for domain adaptation. In particular, CLIP can extract representative target-domain styles directly from textual prompts, enabling adaptation without target images.  Building on CLIP’s language embeddings, Prompt-driven Zero-shot Domain Adaptation (PØDA) generates target-domain visual features from textual descriptions of target conditions and adapts the model accordingly, eliminating the need for target data and framing adaptation as a zero-shot task. A major limitation of PØDA is its dependence on prior knowledge, namely the domain \textit{ID}, to choose the correct domain-specific model. Each target domain requires a separate model guided by CLIP text embeddings, and during inference, the system must first identify the target domain before selecting the appropriate model. This reliance on domain-specific customization reduces flexibility and limits adaptation to diverse or changing environments.

To generalize across multiple target domains without explicit domain identifiers, ULDA~\cite{yang2024unified} trains a unified model and introduces a Text-Driven Rectifier (TDR) to mitigate misalignment between simulated and real target features. Similarly, PØDA and ULDA leverage CLIP text embeddings without accessing target images, but rely on frozen encoders and do not fine-tune for downstream tasks. Although CLIP provides strong global representations, its image-level supervision is insufficient for pixel-level segmentation~\cite{zhou2022extract, ding2022decoupling}, which requires both fine-grained detail and global context~\cite{wang2023coupling, yu2023zero, saire2022global}. Without localized supervision, feature–text misalignment may lead to noisy predictions~\cite{lafon2024gallop, zhou2022extract}. While fine-tuning the vision encoder can improve adaptation to domain-specific variations, it must be carefully designed to preserve out-of-domain generalization under distribution shifts~\cite{kumar2022fine}. Recent work such as ProLIP~\cite{fahes2024fine} highlights the benefits of well-structured fine-tuning, though primarily for classification~\cite{fahes2024simple}, and PODA shows that making CLIP encoders trainable for segmentation is non-trivial and can degrade performance. These findings emphasize the need for carefully designed fine-tuning strategies for robust semantic segmentation.

Full VLM fine-tuning is challenging with limited annotated data \cite{lee2017going,he2024prompting}, motivating parameter-efficient prompt tuning for effective multi-domain adaptation \cite{munir2024exploring,bar2022visual,chen2022adaptformer}.

Prompt learning has been investigated across visual, textual, and multi-modal settings. In textual scenarios, it typically augments inputs with learnable context vectors \cite{zhou2022learning}, while multi-modal prompt tuning jointly optimizes textual and visual prompts within vision–language models to encourage improved cross-modal alignment \cite{zhou2022learning}.
Despite promising progress, these approaches are primarily developed for transformer-based architectures and are less explored for convolutional backbones such as ResNets. Furthermore, while leveraging descriptive prompts generated by large vision-language models (VLMs) can provide useful semantic guidance, such descriptions may not always capture the fine-grained details required for effective feature adaptation in zero-shot domain transfer. This suggests opportunities for developing refined fine-tuning frameworks that better align domain-specific features. Current zero-shot domain adaptation methods continue to face challenges in modeling subtle variations across diverse target domains.

To tackle these challenges, we propose CROMPT (Cross-Modal Prompt Learning for Domain Adaptation), a multi-modal image–text prompt learning framework incorporating progressive domain adaptation. In the zero-shot feature augmentation stage, we introduce Cross-Modal Feature Augmentation (CROMA), which fine-tunes the CLIP vision encoder by integrating Random Visual Prompting (RVP), backbone adaptation, and Hybrid Textual Prompting (HTP) that combines learnable and hard-coded text prompts. By jointly processing prompts from multiple domains through the CLIP text encoder, CROMA captures a variety of candidate styles in each iteration, encouraging style-invariant representations while enriching visual features with complementary semantic cues.
To reduce feature interference when adapting simultaneously to easy and hard domains, we further propose two variants: CROMA-VL, which leverages a vision–language model for domain-conditioned textual prompts, and CROMPT, a sequential easy-to-hard adaptation strategy that progressively refines domain-invariant features. This structured, staged approach enables stable end-to-end training of the baseline segmentation model with enhanced mined features, effectively leveraging rich pre-trained representations for robust fine-tuning under significant domain shifts. The key contributions of this paper are:

\begin{itemize}
    \item We present CROMPT, a prompt-learning framework designed for zero-shot domain adaptation. It incorporates Cross-Modal Feature Augmentation (CROMA) to fine-tune the CLIP vision encoder by combining novel Random Visual Prompting (RVP), backbone adaptation, and Hybrid Textual Prompting (HTP), which unifies learnable and hard-coded text prompts to achieve robust feature alignment across unseen domains.
    
    \item We introduce two adaptation strategies: CROMA-VL, which leverages a vision–language model to generate domain-conditioned prompts for cross-modal feature augmentation, and CROMPT, a sequential easy-to-hard framework that progressively refines domain-invariant features through a two-stage multimodal prompt learning scheme without relying on a VLM.
    
    \item Experiments on two benchmark datasets show that CROMPT surpasses existing state-of-the-art methods while maintaining the baseline model’s computational efficiency, achieving particularly strong performance on challenging domains such as night and rain.
\end{itemize}

\section{Related works}

\textbf{Domain adaptation:} UDA transfers knowledge from labeled source to unlabeled target domains, while OSUDA adapts from a single target image. GDA reduces large domain shifts via intermediate domains but typically requires target data or synthetic domains~\cite{azuma2024zodi}. In contrast, ZSDA requires no target data, leveraging textual descriptions. Methods such as P\O DA~\cite{fahes2023poda} and ULDA~\cite{yang2024unified} exploit CLIP-based semantic guidance for zero-shot adaptation, while recent approaches combine visual alignment with language-guided modulation at the cost of additional memory~\cite{li2025generalized}. These limitations motivate more efficient prompt-based adaptation.
 \\
\textbf{CLIP Fine-Tuning for Downstream Tasks:}  
Although CLIP provides highly transferable representations, adapting it for semantic segmentation requires capturing fine-grained semantics. Methods such as P\O DA enable zero-shot segmentation with frozen CLIP encoders, while DGSS~\cite{fahes2024simple} and ProLIP~\cite{fahes2024fine} show that lightweight adaptation of limited network components is more effective than extensive fine-tuning~\cite{saito2023mind}. However, existing parameter-efficient approaches, including prompt tuning~\cite{chen2022plot, zhou2022learning, zhu2023prompt} and adapters~\cite{zhang2022tip, gao2024clip}, are primarily designed for transformer architectures, incur additional computational overhead, and do not generalize well to convolutional backbones~\cite{fahes2024fine}.

\begin{figure}[t]
    \centering
    \includegraphics[width=85mm, height=40mm]{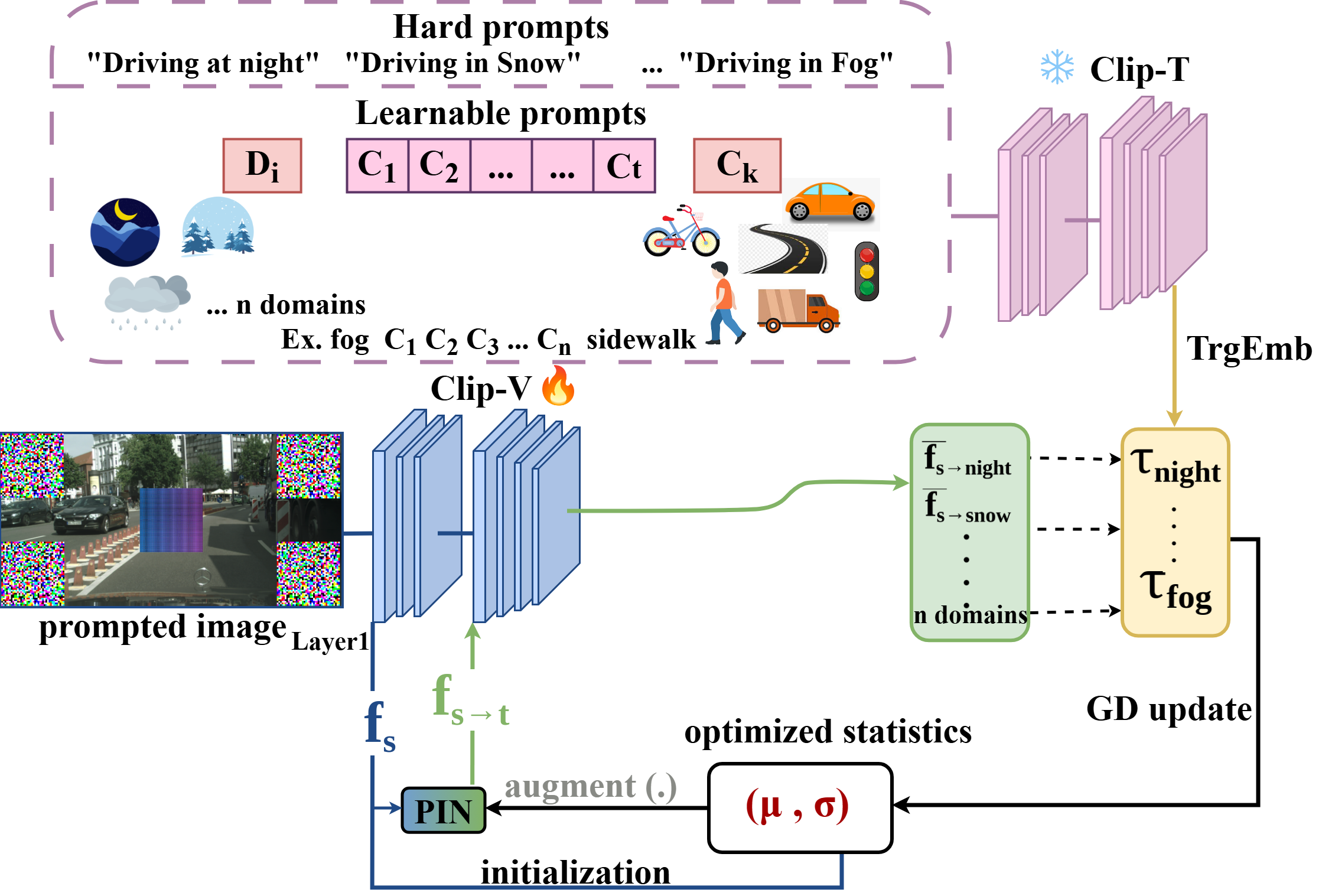}
    \caption{Overview of the proposed zero-shot feature augmentation framework. The source image is processed by the CLIP image encoder (V) to produce target-style features, which are simultaneously aligned with domain-specific target embeddings (e.g., $\tau_{night}, \tau_{rain}$). RVP adds prompts directly to the image, while HTP combines hard prompts with learnable textual prompts for the CLIP text encoder to guide domain-aware feature adaptation.}
    \label{fig:pin}
\end{figure}

\section{Preliminaries}

We review the zero-shot adaptation methods PODA~\cite{fahes2023poda} and ULDA \cite{yang2024unified}, which use textual prompts (\textit{TrgPrompt}) to adapt a segmentation model \textit{M}. The model \textit{M} has a feature encoder $M_{\text{feat}}$ and a prediction module with a decoder and classification head $M_{\text{cls}}$, i.e., $M = (M_{\text{feat}}, M_{\text{cls}})$. Both methods work in two stages: Zero-shot Feature Augmentation, which generates target-domain features using $M_{\text{feat}}$, and Model Adaptation, which updates the decoder with these features and $M_{\text{cls}}$.  
The CLIP image encoder $E_{\text{img}}$ uses a ResNet-50 backbone, and textual prompts are encoded by the CLIP text encoder $E_{\text{text}}$. The target-domain embedding, \textit{TrgEmb}, is computed as $\textit{TrgEmb} = E_{\text{text}}(\text{TrgPrompt})$.\\
\textbf{Zero-shot Feature Augmentation}  proposes Prompt-driven Instance Normalization (PIN) \cite{fahes2023poda} (Eq. \ref{eq_1}), where learnable parameters $\mu$ and $\sigma$, conditioned on a text prompt, capture target domain style by deriving statistics from the hypothetical target feature map $f_{t}$, given by, 
\begin{equation}
    f_{s \rightarrow t} = PIN(f_{s},\mu,\sigma) = \sigma ({\dfrac{f_{s}-\mu(f_{s})}{\sigma(f_{s})}}) + \mu 
    \label{eq_1}
\end{equation} 

here, $\mu(f_{s})$ and $\sigma(f_{s})$ represent the mean and standard deviation of the source features $f_{s}$. PIN maps source features to the target domain $f_{s \rightarrow t}$, which are then projected into the CLIP latent space as $\bar{f}_{s \rightarrow t}$. For style mining, $(\mu, \sigma)$ are optimized via the loss in Eq. (\ref{eq_2}) to maximize the similarity between $\bar{f}_{s \rightarrow t}$ and the CLIP text embeddings \textit{TrgEmb}:
 
\begin{equation}
    L_{\mu,\sigma} (\bar{f}_{s \rightarrow t}, TrgEmb)  = 1 -  {\dfrac{\bar{f}_{s \rightarrow t}.TrgEmb}{\parallel \bar{f}_{s \rightarrow t}\parallel\parallel TrgEmb \parallel}}
    \label{eq_2}
\end{equation}
This ensures alignment between transformed features and text embeddings, facilitating source-to-target adaptation. To maintain the CLIP latent space, $M_{feat}$ remains frozen in PODA. \\
\textbf{Model Adaptation} PODA and ULDA enhance domain adaptation by fine-tuning the segmentation head using only source-domain images together with the mined target-style statistics $(\mu, \sigma)$ corresponding to a specific domain. This process builds upon the base segmentation model \textit{M}, which was originally trained on the source-domain dataset. The base model is further fine-tuned using these extracted style features to improve generalization to previously unseen target domains. The synthesized target-domain features are subsequently employed to train the segmentation head and the classifier $M_{\text{cls}}$. The resulting adapted segmentation model, updated with the mined target-domain characteristics, is denoted as $M_{adp}$.

Differences between simulated and real target-domain features can cause misalignment in the segmentation head, reducing performance. To address this, ULDA introduces a \textit{text-driven rectifier (TDR)} in stage 2, which adjusts features using projected CLIP text embeddings:

\begin{equation}
    \tilde{f}_{s \rightarrow t} = \beta \Big( \tilde{\sigma} \Big( \frac{\bar{f}_{s \rightarrow t} - \mu(\bar{f}_{s \rightarrow t})}{\sigma(\bar{f}_{s \rightarrow t})} \Big) + \tilde{\mu} \Big) + \bar{f}_{s \rightarrow t},
\end{equation}

where $\beta$ is learnable, and $\tilde{\sigma}$ and $\tilde{\mu}$ encode target-domain characteristics. The rectified features $\tilde{f}_{s \rightarrow t}$ are used to fine-tune the segmentation head, preserving domain-specific information and enhancing generalization, with parameters optimized solely through the segmentation loss.

\section{Problem Formulation}

In zero-shot domain adaptation, the goal is to enable a model to perform effectively on unseen target domains without accessing any samples from those domains, i.e., $\{D_{t1}, D_{t2}, \dots, D_{tn}\}$. A model $M$, trained on a source domain $D_s$ for semantic segmentation, leverages textual descriptions (\textit{TrgPrompt}) to anticipate potential target-domain conditions and is adapted into $M_{\text{adp}}$. Our goal is to develop a unified model that generalizes across diverse domain shifts, eliminating the need for explicit domain identifiers during inference.

\section{Cross Modal prompt learning for progressive Domain Adaptation (CROMPT)}

We present a prompt-learning framework designed to enhance cross-modal representations during both feature augmentation and model adaptation. Following the two-stage approach of PØDA and ULDA, our method consists of zero-shot feature augmentation followed by model adaptation. In the zero-shot augmentation stage, we introduce Cross-Modal feature Augmentation (CROMA), which combines Random Visual Prompts (RVP) with Hybrid Textual Prompting (HTP). The framework offers two variants: CROMA-VL, which leverages vision language (VL) models for domain-conditioned textual prompts, and CROMPT, a progressive adaptation strategy tailored for settings where VLMs are not available.

\subsection{Zero-shot Feature Augmentation}

%\subsection{}

%\textbf{Cross Modal feature Augmentation (CROMA):}
To fine-tune the CLIP vision encoder $E_{\text{img}}$ and capture target-domain-representative features, we propose  Cross Modal feature Augmentation (CROMA). This framework combines \textit{Random Visual Prompts (RVP)} for the image encoder with \textit{Hybrid Textual Prompting (HTP)}, integrating both learnable and hard text prompts to enable robust zero-shot feature adaptation across unseen domains. During training, the CLIP image encoder $E_{\text{img}}$ is made trainable, while the CLIP text encoder remains frozen. \\
\textit{(i) Random Visual Prompts (RVP):} We propose a novel visual prompting approach to fine-tune CLIP vision encoders through input-level modifications, inspired by Visual Prompting (VP)~\cite{bahng2022exploring}. In standard VP, continuous, task-specific prompts are added to the input to create a prompted image, while the encoder remains frozen and only the prompt parameters are optimized. In contrast, our Random Visual Prompts (RVPs) are non-learnable and act as noise, while the CLIP vision encoder is jointly fine-tuned alongside these input modifications during training.
Specifically, we use fixed-size patch prompts of size $30 \times 30$. Four corner patches are placed at the edges of the image, and a separate patch is positioned at the center. Let $x \in \mathbb{R}^{3 \times H \times W}$ denote the input image. We define the prompt tensor $p \in \mathbb{R}^{3 \times 30 \times 30}$ as the spatial composition of the five patches:
\begin{equation}
p = p_c^{(tl)} + p_c^{(tr)} + p_c^{(bl)} + p_c^{(br)} + p_m,
\end{equation}
where $p_c^{(\cdot)}$ correspond to the four corner patches (top-left, top-right, bottom-left, bottom-right), and $p_m$ is the center patch with parameter $\theta_m$. All patches occupy only their designated $30 \times 30$ regions, with zero values elsewhere in the image.
The prompted image is then computed as:
\begin{equation}
x' = x + p,
\end{equation}
where the addition is performed element-wise, as shown in Fig. \ref{fig:pin}. This visual prompting strategy is applied exclusively during training and does not increase the number of trainable parameters, since the prompts themselves are non-learnable, while still enabling effective fine-tuning of the CLIP vision encoder.\\
\textit{(ii) Hybrid Textual Prompting (HTP):} 
CLIP employs a vision encoder $\textit{V}$ and a text encoder $\textit{T}$ to generate visual and textual embeddings, which are projected into a shared latent space. During style mining, Eq. (2) aligns these embeddings by updating the normalization parameters $(\mu, \sigma)$ to minimize the gap between the transformed features and their corresponding \textit{TrgEmb}. Unlike PØDA~\cite{fahes2023poda}, which aligns each target style individually, our method performs simultaneous alignment across all styles in a single forward pass (Fig. \ref{fig:pin}). By processing textual prompts from multiple domains, $D_1$ to $D_6$, jointly through the CLIP text encoder, the model considers several candidate styles in each iteration, enabling the learning of style-invariant representations while enriching visual features with complementary semantic cues from the text embeddings.

We introduce a strategy that combines \textit{learnable textual prompts} with \textit{hard-coded prompts} within the training framework. Hard prompts explicitly describe domain conditions (e.g., ``driving at night'' for nighttime scenes), providing stable semantic guidance. For learnable prompts, inspired by CoOp~\cite{zhou2022learning}, we replace selected portions of a prompt template with trainable token embeddings, while preserving the domain identifier ($D_i$) and class label [CLASS] tokens.
This design maintains domain- and category-level semantics, while enabling adaptive learning of contextual representations. For learnable prompts, a textual description such as “driving at night” is split into a target domain token $D_i$ and a class token, where [CLASS] corresponds to the label of each class derived directly from the dataset.
The prompts is tokenized into a sequence of embeddings:
\begin{equation}
\tilde{Y} = \{t_{\text{SOS}}, d_i, t_1, t_2, \dots, t_T, c_k, t_{\text{EOS}}\},
\end{equation}
where $d_i$ represents the target domain embedding, $c_k$ is the embedding of the class label, and $t_{\text{SOS}}$, $t_{\text{EOS}}$ denote the start and end tokens, respectively. To enable adaptive prompt learning, we replace the contextual tokens with learnable vectors $\mathbf{C} = \{C_1, C_2, \dots, C_T\}$. The resulting learnable prompt formulation becomes:
\begin{equation}
\tilde{Y} = \{t_{\text{SOS}}, d_i, C_1, C_2, \dots, C_T, c_k, t_{\text{EOS}}\},
\end{equation}
where $C_1, \dots, C_T$ are trainable context embeddings, while $d_i$ and $c_k$ remain fixed. The resulting prompts are fed into the CLIP text encoder. In Hybrid Textual Prompting (HTP), hard prompts are applied to three domains $\{D_1, D_2, D_3\}$, while soft learnable prompts target three other domains $\{D_4, D_5, D_6\}$, forming mutually exclusive sets. These prompts are combined and processed jointly by the CLIP text encoder in each training iteration. \\
\subsection{Model Adaptation} Discrepancies between synthesized and target-domain features can misalign the segmentation head, degrading performance. Following ULDA~\cite{yang2024unified}, we fine-tune a source-pretrained segmentation model $M$ using the synthesized target-domain features from the feature augmentation stage. The adapted segmentation head and classifier $M_{\text{cls}}$ produce a model $M_{\text{adp}}$ that better captures target-domain characteristics and generalizes to unseen domains.\\
\subsection{Overall loss function} Our approach introduces no additional task-specific loss functions, as the multi-modal prompting mechanism is learned end-to-end. During zero-shot feature augmentation, the loss $\mathcal{L}$ (Eq.~2) aligns the transformed features $\bar{f}_{s \rightarrow t}$ with target text embeddings while preserving category semantics. To further enhance feature adaptation, we incorporate the regional alignment loss $L_r$ and domain consistency loss $L_{DC}$ from ULDA~\cite{yang2024unified}. Once target-domain styles are mined, we adopt ULDA's Stage-2 Text-Driven Rectifier (TDR) and fine-tune the segmentation head using cross-entropy loss, yielding an adapted model $M_{\text{adp}}$ that generalizes to unseen domains without increasing inference complexity.

\begin{figure}[t]
    \centering
    \includegraphics[width=85mm, height=55mm]{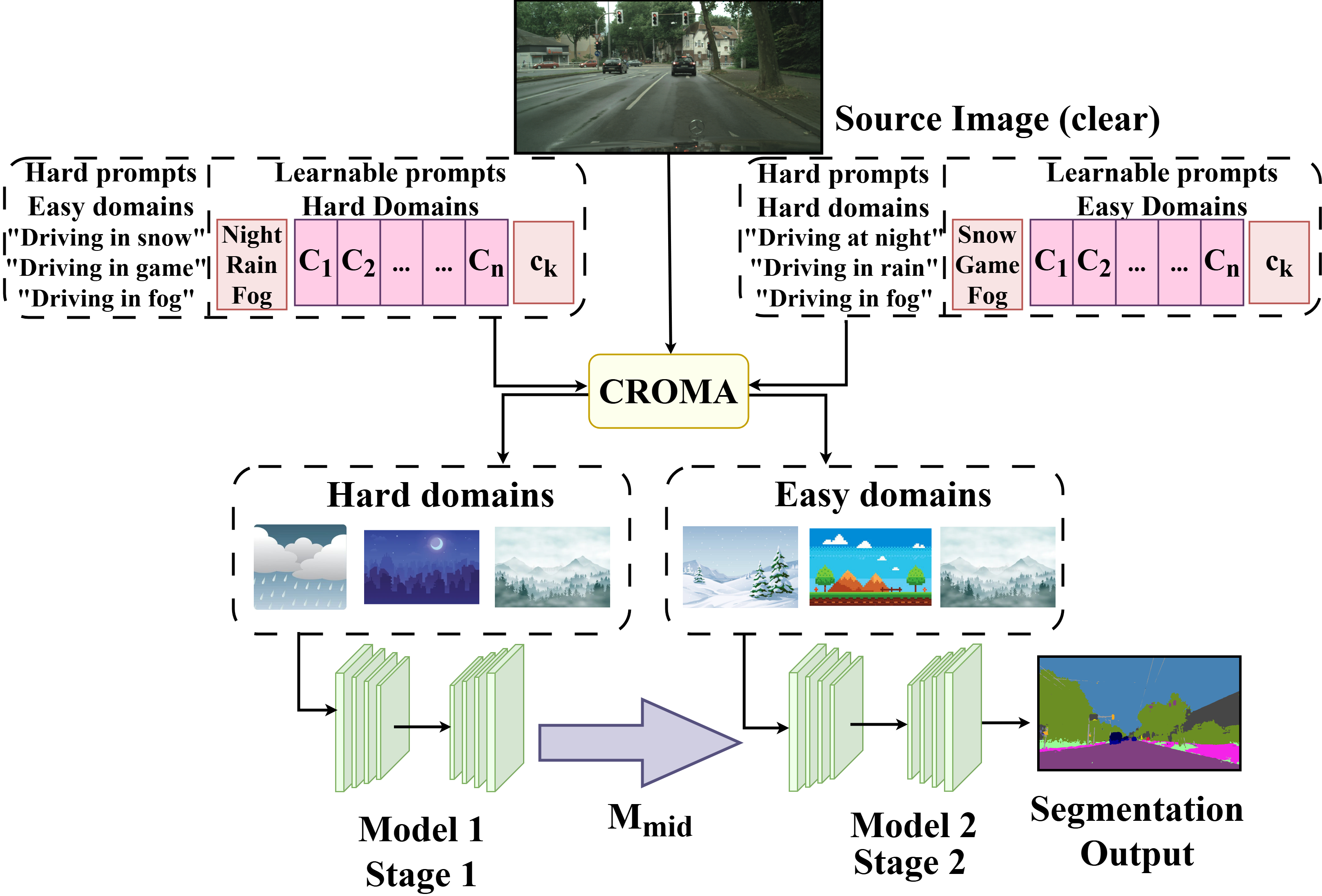}
    \caption{Model adaptation with simulated features using CROMPT. Step 1 applies learnable prompts to hard domains and hard-coded prompts to easy domains, producing the intermediate model $M_{mid}$. In Step 2, the prompt assignments are swapped, learnable prompts are applied to easy domains and hard-coded prompts to hard domains, and the intermediate model $M_{mid}$ is used to initialize this stage, producing the final adapted model $M_{adp}$.}
    \label{fig:model_adapt}
\end{figure}

\subsection{ Cross-Modal Adaptation}

We propose two frameworks for cross-modal adaptation. \\
\textit{(i) Cross Modal feature Augmentation using Vision language model (CROMA-VL):} In this framework, the QWEN-VL \cite{bai2025qwen3} model is used for zero-shot feature augmentation. Learnable prompts are applied to easier domains (Snow, Game, Fog), while hard-coded prompts target more challenging domains (Rain, Night, Fog), with Fog serving as an intermediate-complexity domain. For the hard prompts, source images are passed through QWEN-VL, conditioned on specific environmental contexts such as rainy scenes, to generate domain-adapted textual descriptions. For example, a rain-conditioned prompt generated by QWEN-VL would be:
\textit{"The image shows a rainy day in a city, with cars parked along the side of the road. There are multiple cars, including a silver car parked in front of a black car. A person is walking down the street, and a bicycle is parked nearby. The weather conditions are rainy, and the lighting conditions are dim, with shadows visible."}
These domain-conditioned textual descriptions, together with the Hybrid Textual Prompting (HTP) strategy and visual prompts, are then used during the zero-shot feature augmentation stage, followed by model adaptation.\\
\textit{(ii) Cross Modal prompt learning for progressive Domain Adaptation (CROMPT):}
When the use of large vision language models is restricted, for instance due to privacy concerns, we propose a two-step progressive adaptation strategy that leverages multi-modal prompts.
In the first step, learnable prompts are applied to the challenging domains (Night, Rain, Fog), while hard-coded prompts are assigned to more easier domains (Snow, Game, Fog). CROMA extracts target-domain-representative features, with domain difficulty determined by factors such as distribution shift, visual complexity, or semantic divergence from the source. These features are then used to fine-tune the segmentation head and image encoder following ULDA’s text-driven rectifier (TDR) strategy, resulting in an intermediate adapted model $M_{\text{mid}}$.\\
In the second step, the roles of the prompts are reversed: learnable prompts target the easy domains, and hard-coded prompts target the harder domains. Feature augmentation and model adaptation are performed again, initializing from $M_{\text{mid}}$, which facilitates knowledge transfer from simpler to more complex domains and ultimately produces the final adapted model $M_{\text{adp}}$, as shown in Fig. \ref{fig:model_adapt}.
This progressive, structured approach enables the model to incrementally learn domain-invariant representations, preserving robustness from previously seen domains while aligning both the segmentation head and encoder to increasingly challenging target styles. The resulting model $M_{\text{adp}}$ generalizes effectively across unseen domains without requiring prior domain information, leveraging variations from both easier and harder domains to achieve strong zero-shot performance. 
For model complexity, CROMA introduces only a minimal number of learnable parameters from the textual prompts, while the random visual prompts do not add any additional parameters. Importantly, both adaptation steps maintain the same computational cost as the baseline model during inference.

\begin{figure}[t]
    \centering
    \includegraphics[width=85mm, height=55mm]{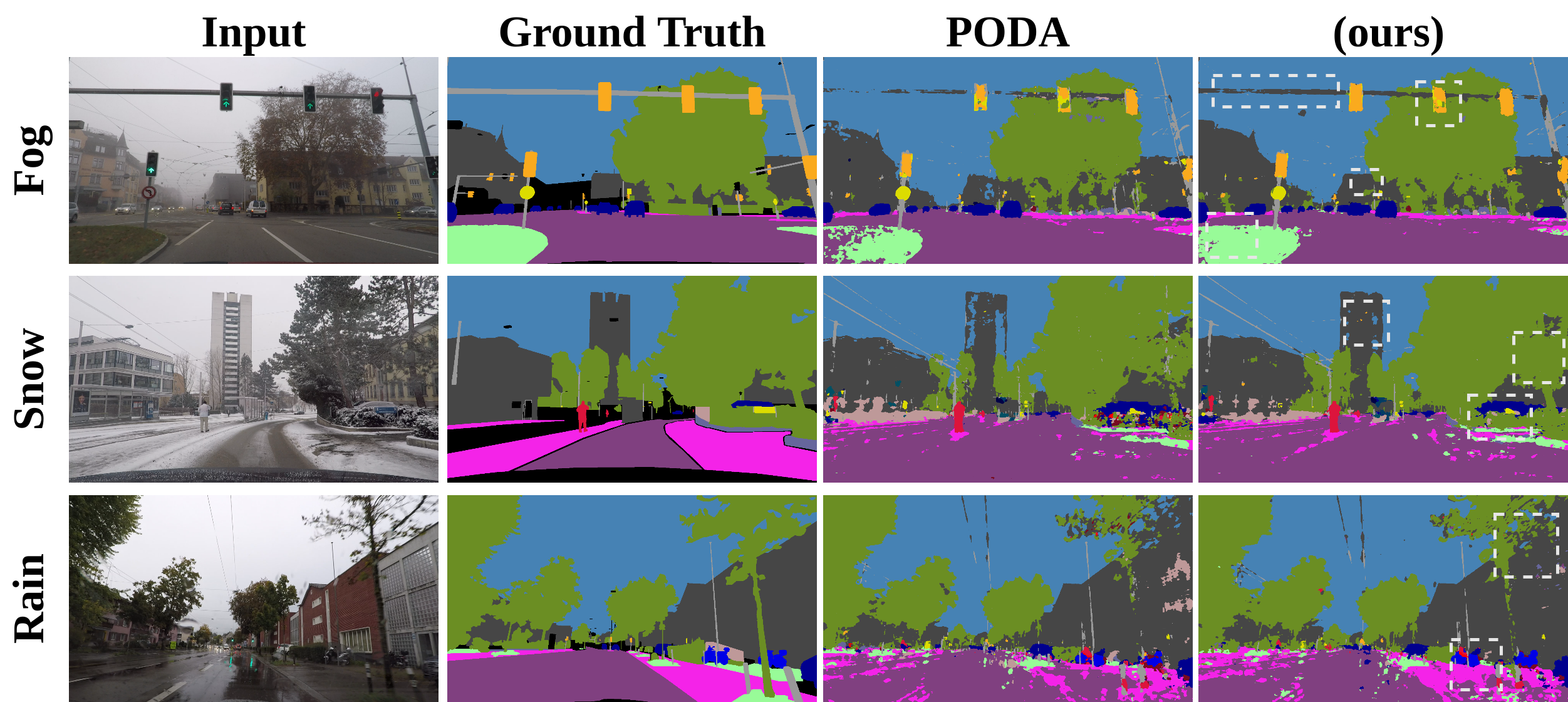}
    \caption{Comparison of output predictions under adverse weather conditions between PODA and our model. Our method produces more accurate and consistent segmentation across challenging scenarios.}
    \label{fig:pred}
\end{figure}

\section{Experimental Results}

\subsection{Experimental Setup}
\textbf{Datasets and Evaluation}  In our experimental setup, Cityscapes is used as the source domain, containing 2,975 training images and 500 validation images. The trained models are evaluated on the target datasets ACDC \cite{sakaridis2021acdc} and GTA5 \cite{richter2016playing}. Evaluation on ACDC follows its official 406-image validation split, while GTA5 is assessed on a 1,000-image subset in line with established protocols \cite{yang2024unified,fahes2023poda}. All datasets share a unified 19-class label space. During testing, target images are processed at their native resolutions, and results are reported in terms of mIoU (\%).\footnote{Reported results have been reproduced using open-source code.} \\
\textbf{Implementation Details.} We employ DeepLabv3+ \cite{chen2018encoder} as the baseline segmentation network, treating it as the source model, and initialize it with a CLIP-ResNet-50 backbone \cite{radford2021learning}. The network is first fully trained on the source dataset following the protocol of \cite{fahes2023poda}.
For zero-shot feature enhancement, the PIN module is trained under the PODA framework \cite{fahes2023poda} with a mini-batch size set to four samples per iteration. During the subsequent adaptation stage, the pre-trained source model is further refined using the augmented PIN representations by updating the classification head $M_{cls}$ for 2000 iterations, again operating with four samples per mini-batch, following \cite{fahes2023poda}. 

\begin{table}[]
\caption{Performance comparison (mIoU \%) for zero-shot domain adaptation. Cityscapes is used as the source domain, with ACDC and GTA5 as target domains. * denotes results reproduced from official code, and \textit{TrgPrompt} refers to the target prompt.}
\centering
\begin{tabular}{cccc}
\specialrule{.1em}{.05em}{.05em} 
\rowcolor[HTML]{C0C0C0} 
\textbf{Source} & \textbf{Target Eval.} & \textbf{Method}                                                                                  & \textbf{mIoU}                                                                       \\ \specialrule{.1em}{.05em}{.05em} 
\multicolumn{4}{c}{\cellcolor[HTML]{EFEFEF}\textit{TrgPrompt} = "driving at night"}                                                                                                                   \\
CS  \quad $\mapsto$   & ACDC - Night & \begin{tabular}[c]{@{}c@{}}source-only\\ CLIPStyler\\ PODA*\\ ULDA*\\ CROMA-VL \\CROMPT \end{tabular} & \begin{tabular}[c]{@{}c@{}}18.31\\ 21.38\\ \textbf{23.97}\\ \underline{21.21}\\ 23.91 \\23.30\end{tabular}     \\
\multicolumn{4}{c}{\cellcolor[HTML]{EFEFEF}\textit{TrgPrompt} = "driving in snow"}                                                                                                                    \\
CS  \quad $\mapsto$    & ACDC - Snow  & \begin{tabular}[c]{@{}c@{}}source-only\\ CLIPStyler\\ PODA*\\ ULDA*\\ CROMA-VL \\ CROMPT\end{tabular} & \begin{tabular}[c]{@{}c@{}}39.28\\ 41.09\\ 40.59\\ \underline{43.01}\\ \textbf{44.09}\\ {44.04}\end{tabular}  \\
\multicolumn{4}{c}{\cellcolor[HTML]{EFEFEF}\textit{TrgPrompt} = "driving under rain"}                                                                                                                    \\
CS  \quad $\mapsto$    & ACDC - Rain  & \begin{tabular}[c]{@{}c@{}}source-only\\ CLIPStyler\\ PODA*\\ ULDA*\\ CROMA-VL \\CROMPT\end{tabular} & \begin{tabular}[c]{@{}c@{}}38.20\\ 37.17\\ 39.20\\ \underline{41.96}\\ {43.69} \\ \textbf{44.41}\end{tabular}  \\
\multicolumn{4}{c}{\cellcolor[HTML]{EFEFEF}\textit{TrgPrompt} = "driving in fog"}                                                                                                                     \\
CS   \quad $\mapsto$   & ACDC - Fog   & \begin{tabular}[c]{@{}c@{}}PODA*\\ ULDA*\\ CROMA-VL \\CROMPT \end{tabular}                            & \begin{tabular}[c]{@{}c@{}}49.34\\ \underline{51.53}\\ {51.54} \\ \textbf{51.88}\end{tabular}                  \\
\multicolumn{4}{c}{\cellcolor[HTML]{EFEFEF}\textit{TrgPrompt} = "driving in a  game"}                                                                                                                    \\
CS  \quad $\mapsto$    & GTA5         & \begin{tabular}[c]{@{}c@{}}source-only\\ CLIPStyler\\ PODA*\\ ULDA*\\ CROMA-VL \\CROMPT\end{tabular} & \begin{tabular}[c]{@{}c@{}}39.59\\ 38.73\\ 40.91 \\ \underline{40.93}\\ \textbf{41.60}\\ \textbf{41.84}\end{tabular} \\ \specialrule{.1em}{.05em}{.05em} 
\end{tabular}

\label{Table:acdc}
\end{table}

\begin{table}[t]
\caption{Ablation study showing the incremental contribution of CROMA, CROMA-VL, and CROMPT on top of ULDA.}
\centering
\begin{tabular}{lcccc}
\hline
\rowcolor[HTML]{C0C0C0} 
\textbf{Method} & \textbf{CROMA} & \textbf{CROMA-VL} & \textbf{CROMPT} & \textbf{\begin{tabular}[c]{@{}c@{}}Mean \\ (mIoU)\end{tabular}} \\
\hline

ULDA &  &  &  & 39.25 \\

ULDA  & \ding{51} &  &  & 40.69 \\

ULDA & \ding{51} & \ding{51} &  & \underline{40.81} \\

ULDA & \ding{51} & \ding{51} & \ding{51} & \textbf{40.88} \\
ULDA & \ding{51} &  & \ding{51} & \textbf{40.91} \\

\hline
\end{tabular}
\label{ablation}
\end{table}

\begin{table}[]
\caption{Segmentation performance (mIoU \%) of source and adapted models, with $\Delta$ indicating gains over the source-only baseline. SM-PPM uses one unlabeled target image per domain for adaptation, whereas CROMPT operates in a fully zero-shot manner without target data.}
\centering
\begin{tabular}{
>{\columncolor[HTML]{FFFFFF}}c 
>{\columncolor[HTML]{FFFFFF}}c 
>{\columncolor[HTML]{FFFFFF}}c 
>{\columncolor[HTML]{FFFFFF}}c }
\specialrule{.1em}{.05em}{.05em}
\rowcolor[HTML]{C0C0C0} 
\textbf{Source} & \textbf{\begin{tabular}[c]{@{}c@{}}Target \\ Eval.\end{tabular}} & \textbf{\begin{tabular}[c]{@{}c@{}}One-shot \\ SM-PPM\end{tabular}} & \textbf{\begin{tabular}[c]{@{}c@{}}Zero-shot \\ CROMPT\end{tabular}} \\ \hline
CS \quad    $\mapsto$         & \begin{tabular}[c]{@{}c@{}}ACDC Night\end{tabular}            & \begin{tabular}[c]{@{}c@{}}13.07 / 14.60 \\ ($\Delta$=1.53)\end{tabular}   & \begin{tabular}[c]{@{}c@{}}18.31 / 23.29 \\ ($\Delta$=\textbf{4.98})\end{tabular}  \\
CS  \quad     $\mapsto$       & \begin{tabular}[c]{@{}c@{}}ACDC Snow\end{tabular}             & \begin{tabular}[c]{@{}c@{}}32.60 / 35.61 \\ ($\Delta$=3.01)\end{tabular}   & \begin{tabular}[c]{@{}c@{}}39.28 / 44.04 \\ ($\Delta$=\textbf{4.76})\end{tabular}                                                                   \\
CS    \quad  $\mapsto$        & \begin{tabular}[c]{@{}c@{}}ACDC  Rain\end{tabular}             & \begin{tabular}[c]{@{}c@{}}29.78 / 32.23 \\ ($\Delta$=2.45)\end{tabular}                & \begin{tabular}[c]{@{}c@{}}38.20 / 44.41 \\ ($\Delta$=\textbf{6.21})\end{tabular}                                                               \\ \specialrule{.1em}{.05em}{.05em}
\end{tabular}

\label{Table:OSUDA}
\end{table}

\subsection{Quantitative Comparison}

\textbf{Comparison with 0-shot adaptation techniques:}  Table~\ref{Table:acdc} compares recent state-of-the-art methods on ACDC and GTA5, including the source-only baseline, CLIPstyler~\cite{kwon2022clipstyler}, PODA~\cite{fahes2023poda}, ULDA~\cite{yang2024unified}, and our approaches, CROMA-VL and CROMPT. While CLIPstyler performs text-guided style transfer, its stylization artifacts often limit segmentation performance. On ACDC, CROMPT outperforms PODA and ULDA by \textbf{2.63\%} and \textbf{1.63\%} mIoU on average, respectively. Similar gains are observed on GTA5, with improvements of \textbf{0.93} and \textbf{0.91} mIoU over PODA and ULDA. CROMA-VL also consistently surpasses PODA and ULDA by \textbf{2.33\%} and \textbf{1.55\%} on ACDC. Under adverse conditions, CROMA-VL improves over ULDA by \textbf{2.7\%} in source-to-night adaptation, while CROMPT achieves a \textbf{2.45\%} gain in source-to-rain. Although PODA performs competitively in the night setting, it assumes prior knowledge of the target domain, which our framework does not require.\\
\textbf{Comparison to one-shot adaptation technique} CROMPT is compared with the OSUDA method SM-PPM \cite{wu2022style}, which uses a single unlabeled target image per domain to guide style adaptation as shown in Table \ref{Table:OSUDA}. While SM-PPM trains separate models per domain using ResNet-101 and DeepLabv2, CROMPT uses a single zero-shot model with ResNet-50 and DeepLabv3+. Given these differences, we focus on relative gains over each method's source-only baseline. CROMPT achieves higher improvements both in absolute and relative terms, despite requiring no target domain images.\\\textbf{Qualitative results} Figure \ref{fig:pred} presents qualitative results, with training on Cityscapes and validation on the ACDC dataset. 
\subsection{Ablation Study} 

\textbf{Ablation study for visual prompts} Table \ref{vp-ablation} compares learnable visual prompts placed at various locations with RVP under the Hybrid text prompting protocol, for CROMA-VL. The results indicate that RVP consistently delivers the best performance across settings.\\
\textbf{Ablation with key components} Table~\ref{ablation} presents the ablation study of the proposed framework, including a comparison between CROMA-VL and CROMPT. While incorporating Qwen as a vision--language model improves performance, simple hard-coded prompts (e.g., ``driving at \{domain name\}'') achieve the best overall results.\\

\begin{table}
\centering
\setlength{\aboverulesep}{0pt}
\setlength{\belowrulesep}{0pt}
\setlength{\tabcolsep}{6pt}
\caption{Comparison of prompting strategies across Easy (R/S/F) and Hard (R/F/N) domains.}
\resizebox{\linewidth}{!}{%
\begin{tabular}{@{}ccc@{}}
\rowcolor[HTML]{C0C0C0} 
\toprule
\textbf{Easy domains (R/S/F)} & \textbf{Hard domains (R/F/N)} & \textbf{Mean IoU} \\
\midrule
CoOp & CoOp & 39.45 \\

CoOp & Hand-coded Description & 40.43 \\

Hand-coded Description & CoOp & 39.56 \\

Qwen-VL & Qwen-VL & 39.99\\

CoOp & Qwen-VL & 40.81 \\

\bottomrule
\end{tabular}
}
\label{promtping}
\end{table}
\textbf{Ablation with textual prompts} Table \ref{promtping} presents an ablation analysis of different textual prompting strategies across easy and hard domains. The easy domains include Game, Snow, and Fog, while the hard domains comprise Rain, Night, and Fog. The results indicate that employing vanilla CoOp alone or directly applying Qwen-VL does not provide performance gains. The strongest results for CROMA-VL are obtained using a structured strategy, where CoOp is applied to the easy domains and Qwen-VL is used for the harder domains.\\
\textbf{Ablation two stage adaptation} Table \ref{two-stage-ablation} presents an ablation study of the proposed two-stage adaptation framework using different prompting strategies across easy domains, Snow (S), Game (G), and Fog (F) and hard domains, Rain (R), Fog (F), and Night (N). The analysis incorporates Random Visual Prompts (RVP) together with various textual prompting configurations. The hand-coded descriptions explicitly define domain conditions, e.g., “driving at night” for the night domain.
As observed, the structured progressive adaptation strategy employed in CROMPT achieves the best overall performance compared to other prompting combinations. In addition, extending the framework to a three-stage adaptation, using either Qwen-VL or hand-coded descriptions, leads to a consistent performance drop, yielding results close to the ULDA baseline model.

\begin{table}[t]
\centering
\setlength{\aboverulesep}{0pt}
\setlength{\belowrulesep}{0pt}
\setlength{\tabcolsep}{4pt}
\renewcommand{\arraystretch}{1.1}

\caption{Two-stage progressive adaptation analysis comparing Learnable prompts, Hard Prompt Descriptions, and Qwen-VL across easy (S, G, F) and hard (N, R, F) domains.}

\begin{tabular}{@{}ccc@{}}
\toprule
\rowcolor[HTML]{C0C0C0}
\textbf{Stage 1} & \textbf{Stage 2} & \textbf{mIoU} \\
\midrule

\begin{tabular}[c]{@{}c@{}}Learnable: S/G/F \\ Hard Desc.: N/R/F\end{tabular}
&
\begin{tabular}[c]{@{}c@{}}Learnable: N/R/F \\ Hard Desc.: S/G/F\end{tabular}
& 40.19 \\

\midrule

\begin{tabular}[c]{@{}c@{}}Learnable: N/R/F \\ Hard Desc.: S/G/F\end{tabular}
&
\begin{tabular}[c]{@{}c@{}}Learnable: S/G/F \\ Hard Desc.: N/R/F\end{tabular}
& \textbf{40.91} \\

\midrule

\begin{tabular}[c]{@{}c@{}}Learnable: N/R/F \\ Qwen-VL: S/G/F\end{tabular}
&
\begin{tabular}[c]{@{}c@{}}Learnable: S/G/F \\ Qwen-VL: N/R/F\end{tabular}
& 40.76 \\

\midrule

\begin{tabular}[c]{@{}c@{}}Learnable: N/R/F \\ Hard Desc.: S/G/F\end{tabular}
&
\begin{tabular}[c]{@{}c@{}}Learnable: S/G/F \\ Qwen-VL: N/R/F\end{tabular}
& 40.88 \\

\bottomrule
\end{tabular}
\label{two-stage-ablation}
\end{table}

\begin{table}
\centering
\setlength{\aboverulesep}{0pt}
\setlength{\belowrulesep}{0pt}
\setlength{\tabcolsep}{4pt}
\renewcommand{\arraystretch}{1.1}

\caption{Ablation study comparing learnable visual prompts with non-learnable (noise-based) prompts.}

\begin{tabular}{@{}ccc@{}}
\toprule
\rowcolor[HTML]{C0C0C0} 
\textbf{Prompts} & \textbf{Setting} & \begin{tabular}[c]{@{}c@{}}\textbf{ACDC} \\ \textbf{Mean}\end{tabular} \\
\midrule

\begin{tabular}[c]{@{}c@{}}Visual Prompts \\ (Learnable)\end{tabular}
&
\begin{tabular}[c]{@{}c@{}}Padded \\ Random Patch \\ Corner + Middle\end{tabular}
&
\begin{tabular}[c]{@{}c@{}}39.34 \\ 39.65 \\ 39.62\end{tabular}
\\

\midrule

\begin{tabular}[c]{@{}c@{}}Visual Prompts \\ (Not Learnable)\end{tabular}
& Corner + Middle
& 40.81 \\

\bottomrule
\end{tabular}
\label{vp-ablation}
\end{table}

\section{Conclusions}
We presented \textbf{CROMPT}, a multi-modal prompt learning framework for zero-shot domain adaptation in semantic segmentation. By combining \textbf{Cross-Modal Feature Augmentation (CROMA)} with Random Visual Prompting and Hybrid Textual Prompting (HTP), our method effectively aligns visual and textual representations across unseen domains. We also introduced \textbf{CROMA-VL}, which leverages vision--language models for domain-conditioned prompts, and a progressive easy-to-hard adaptation strategy. Experiments on two benchmark datasets show that our approach consistently outperforms existing methods, improving the source model by \textbf{4.62\% mIoU} on average and surpassing \textbf{ULDA} and \textbf{P\O DA} by up to \textbf{1.63\%} and \textbf{2.80\% mIoU}, respectively, while preserving the inference-time complexity of the original segmentation model.

\bibliographystyle{splncs04}
\bibliography{main}
\end{document}